\documentclass[letterpaper, 10 pt, conference]{ieeeconf}  % Comment this line out if you need a4paper

\IEEEoverridecommandlockouts                              % This command is only needed if 
\usepackage{graphics} % for pdf, bitmapped graphics files
\usepackage{epsfig} % for postscript graphics files
\usepackage{times} % assumes new font selection scheme installed
\usepackage{amsmath} % assumes amsmath package installed
\usepackage{amssymb}  % assumes amsmath package installed
\usepackage{subfigure}
\usepackage{booktabs}
\usepackage{cite}
\usepackage{float}
\usepackage{xcolor}
\usepackage{multirow}
\usepackage{boldline}
\usepackage{soul}

\title{\LARGE \bf
LHMap-loc: Cross-Modal Monocular Localization Using LiDAR  Point Cloud Heat Map 
}

\author{Xinrui Wu\textsuperscript{\rm 1*}\qquad~ Jianbo Xu\textsuperscript{\rm 1*}\qquad~ Puyuan Hu\textsuperscript{\rm 1}\qquad~ Guangming Wang\textsuperscript{\rm 2}\qquad~Hesheng Wang\textsuperscript{\rm 1}\qquad~ % <-this % stops a space
\thanks{*The first two authors contributed equally. This work was supported in part by the Natural Science Foundation of China under Grant 62225309, Grant 62073222, Grant U21A20480, and Grant U1913204; in part by the Science and Technology Commission of Shanghai Municipality under Grant 21511101900; and in part by the Open Research Projects of Zhejiang Laboratory under Grant 2022NB0AB01. Corresponding Author: Hesheng Wang. 
}% <-this % stops a space
\thanks{\textsuperscript{\rm 1}X. Wu, J. Xu, P. Hu and H. Wang are with Department of Automation, Key Laboratory of System Control and Information Processing of Ministry of Education, Key Laboratory of Marine Intelligent Equipment and System of Ministry of Education, Shanghai Engineering Research Center of Intelligent Control and Management, Shanghai Jiao Tong University, Shanghai 200240, China.}
\thanks{
\textsuperscript{\rm 2}G. Wang is with Department of Engineering, University of Cambridge, Cambridge CB2 1PZ, U.K. (e-mail: gw462@cam.ac.uk).}%
}

\begin{document}
\maketitle
\thispagestyle{empty}
\pagestyle{empty}
	
%%%%%%%%%%%%%%%%%%%%%%%%%%%%%%%%%%%%%%%%%%%%%%%%%%%%%%%%%%%%%%%%%%%%%%%%%%%%%%%%
\begin{abstract}

Localization using a monocular camera in the pre-built LiDAR point cloud map has drawn increasing attention in the field of autonomous driving and mobile robotics. However, there are still many challenges (e.g. difficulties of map storage, poor localization robustness in large scenes) in accurately and efficiently implementing cross-modal localization. To solve these problems, a novel pipeline termed LHMap-loc is proposed, which achieves accurate and efficient monocular localization in LiDAR maps. Firstly, feature encoding is carried out on the original LiDAR point cloud map by generating offline heat point clouds, by which the size of the original LiDAR map is compressed. Then, an end-to-end online pose regression network is designed based on optical flow estimation and spatial attention to achieve real-time monocular visual localization in a pre-built map. In addition, a series of experiments have been conducted to prove the effectiveness of the proposed method. Our code is available at: https://github.com/IRMVLab/LHMap-loc.
  
\end{abstract}

%%%%%%%%%%%%%%%%%%%%%%%%%%%%%%%%%%%%%%%%%%%%%%%%%%%%%%%%%%%%%%%%%%%%%%%%%%%%%%%%
\section{Introduction}

Localization is a critical technology \cite{caselitz2016monocular} in autonomous driving and robotics that underlies other downstream tasks, such as navigation and control. Map-based localization is often utilized to alleviate the error caused by satellite signal blocking in GNSS localization methods\cite{wang2016robust, zumberge1997precise, zou2013urtk} and the accumulated drift in Simultaneous Location And Mapping (SLAM) methods \cite{xu2022fast,lin2022r,zheng2022fast}. LiDAR is the primary sensor in constructing the map because point clouds are not affected by light changes in the environment. However, LiDAR sensors are expensive and tend to be large in size, while monocular cameras are small and inexpensive, making them easier to be equipped on mobile devices. Therefore, the technology of visual localization in LiDAR maps has broad application prospects.

\begin{figure}[t]
   \centering
   \includegraphics[width=1\linewidth]{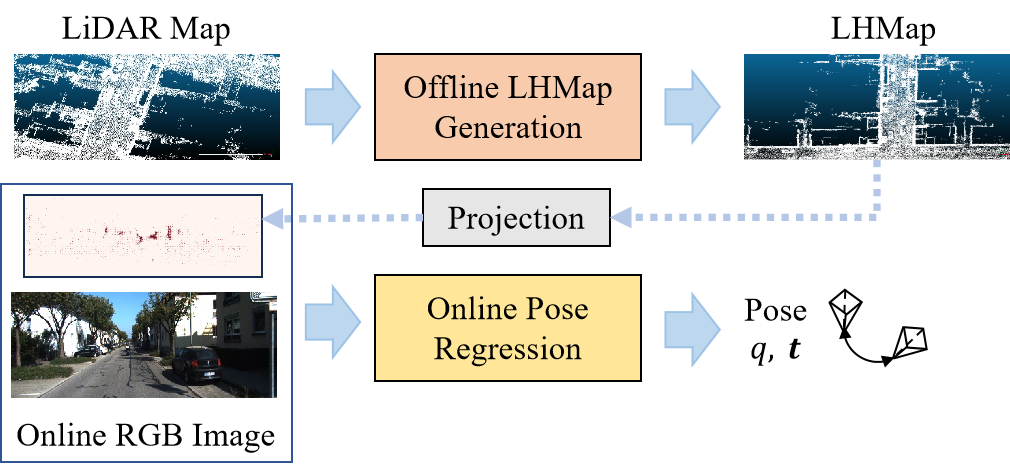}
   \vspace{-0.5cm}
   \caption{Monocular localization pipeline using LiDAR point cloud heat map (LHMap). The pipeline consists of an offline LHMap generation network to build LHMap, and an online pose regression network to achieve real-time localization with pre-built LHMap.}
   \vspace{-0.4cm}
   \label{figure:introduction}
\end{figure}
   
However, there are two main challenges regarding this technology. The first is the large consumption of storage and calculation for LiDAR maps. Since point clouds are  disordered, it usually requires heavy computation to extract features from original LiDAR maps. The second challenge is the 3D-2D cross-modal feature matching. Since the point cloud maps contain 3D coordinate information, while the camera images contain 2D RGB pixels, feature matching cannot be carried out directly between the camera images and the point cloud maps. PnP based methods \cite{PnP, lepetit2009ep} cannot provide reasonable solutions because the correspondence between 3D points and 2D pixel data is unknown. Besides, methods that generate 3D point clouds from 2D images for matching and regression \cite{caselitz2016monocular, Cross-Modal} suffer from depth inaccuracies. Recent HD map based localization methods \cite{wang2021visual,wang2022ltsr,petek2022robust}  rely on specific geometric structures such as lane lines, road signs which may not appear in some rural roads or parks. Additionally, HD map annotations are time-consuming and labour-intensive. Therefore, HD map based methods are not suitable for all occasions.

To deal with the above problems, we propose a monocular localization pipeline termed LHMap-loc as shown in Fig. \ref{figure:introduction}. We refer to the sorted and filtered LiDAR map as LiDAR point cloud Heat Map (LHMap). In this pipeline, LHMap generation is first conducted on the original LiDAR point cloud through offline supervised training, retaining the key features and compressing the point cloud map. Then, the 6 Degree-of-Freedom (DoF) poses are predicted by a pose regression network based on optical flow prediction and spatial attention weighting. In this end-to-end network, real-time high-precision pose regression is realized. The main contributions of this paper are as follows: 

\begin{itemize}
\item 
We propose a monocular visual localization pipeline named LHMap-loc. This pipeline can compress and encode the features of the point cloud map in an offline way, and carry out monocular localization online. The whole pipeline is realized by the deep learning method. 

\item A pose regression algorithm based on optical flow prediction and spatial attention weighting is designed. The algorithm achieves a cross-modal fusion of 3D and 2D features,  enabling end-to-end pose estimation. 
\item Numerous experiments have been conducted on the automatic driving datasets, KITTI \cite{kitti} and Argoverse\cite{argo} datasets. In addition, we conduct real-world experiments on our own wheeled vehicle platform. The results show that the proposed LHMap-loc performs better in terms of precision and efficiency than the state-of-the-art (SOTA) methods\cite{cmrnet,cmrnet++,hyper,i2d}.

\end{itemize}

\section{Related Work}

3D-2D cross-modal localization has always been a challenging task in terms of autonomous driving and robotics. The existing methods can be roughly divided into two categories according to the way of pose regression. One is the traditional methods which construct geometric residuals and calculate pose through mathematical optimization, and the other is the deep learning based methods which encode cross-model feature through MLPs and regress pose from neural networks.   
\subsection{Geometric Optimization Based Localization}
For the problem of monocular visual localization in pre-built maps, 
Caselitz et al. \cite{caselitz2016monocular} utilizes the feature points generated by ORB-SLAM\cite{mur2015orb} to obtain 3D points. Then they utilize Iterative Closest Point (ICP) \cite{ICP} methods to conduct point registration and compute the real-time poses. Yu et al. \cite{yu2020monocular} first extract the line features in the monocular image and point cloud. They then achieve cross-modal localization  by optimizing the distance and angle residuals between line segments. Ye et al.\cite {ye2020monocular,ye20213d} re-build the original point cloud as a Surfel map composed of Surfel descriptors, and then construct the visual-Surfel residual to proceed pose optimization. Similarly, Huang et al. \cite{huang2020gmmloc} model map points using a Gaussian Mixture Model. Recently, Zhang et al. \cite{Cross-Modal} propose to combine semantic point cloud for cross-modal localization. They first construct point cloud maps containing semantic information, and then perform registration and pose regression with the semantic ORB feature points constructed in real time. However, these methods have disadvantages. Firstly, visual features such as feature points and line segments are greatly affected by the lighting condition and are not robust enough, which is easy to cause mismatching. In addition, dynamic objects and irregular noisy points have a crucial impact on Surfel and GMM models during the construction of LiDAR maps.

\subsection{Deep Matching Based  Localization }
Due to the powerful feature coding capability of deep learning networks, Zhou et al. \cite{zhou2020da4ad} encode the features of map points based on image heat maps, and then perform monocular localization through online pose regression. \cite{cmrnet} propose the pose regression method based on 2D optical flow estimation for the first time. Then, Chang et al. \cite{hyper} add LiDAR point cloud feature coding and voxel down-sampling modules on the basis of the original CMRNet\cite{cmrnet}. CMRNet++ \cite{cmrnet++,i2d} adds PnP module following CMRNet\cite{cmrnet}. They estimate the point cloud and pixel matching relationship by 2D optical flow estimation, and then respectively adopt the EPnP \cite{lepetit2009ep} and BPnP\cite{bpnp} method to get the final pose. Miao et al. \cite{PosesAsQueires} propose a novel transformer-based neural network to register images to LiDAR maps, and introduce pose queries to boost the certainty of networks. However, these methods have some problems, such as excessive storage of point cloud maps, poor accuracy and low efficiency of pose regression. 
%------------------------------------------------------------------------

\section{LHMap-loc Method}
\label{section:Proposed Methods}

% 字母和图表的对应关系，针对fig2进行overview的叙述
\begin{figure*}[h]
   \centering
   \includegraphics[width=0.9 \linewidth]{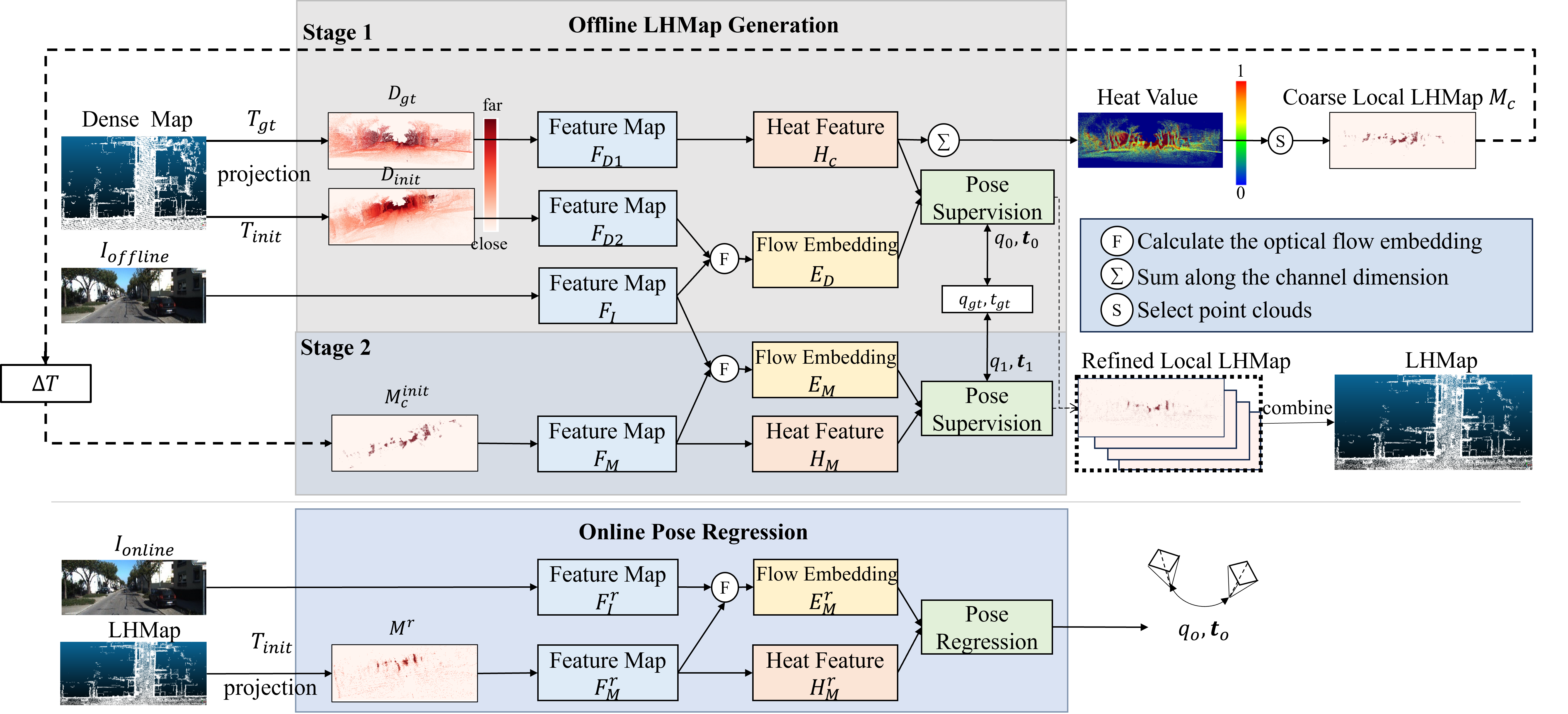}
   \caption{Detailed pipeline of LHMap-loc. It includes the offline LHMap generation network and the online pose regression network. In stage 1 of the offline network, $D_{gt}$ is used to generate heat feature $H_c$, and the coarse local LHMap $M_c$ is selected by the heat value calculated by $H_c$. $D_{init}$ and $I_{offline}$ are used to generate the flow embedding $E_D$. In stage 2, the initial coarse local LHMap $M_c^{init}$ and $I_{offline}$ are used to generate the flow embedding $E_M$ and the heat feature $H_M$. Both $E_D$ and $E_M$ are used for pose supervision by spatial attention weighting. In the online pose regression network, the real-time local LHMap $M_r$ and $I_{online}$ are used to regress the real-time 6-DoF pose.}
   \vspace{-0.2cm}
   \label{figure:pipeline}
    \vspace{-1mm}   
\end{figure*}

% 字母要斜体
\subsection{System Overview} 
%  We denote the RGB image as $I\in{\mathbb{R}^{3\times{H}\times{W}}}$, where $H$ and $W$ represent the height and width of the image respectively. We denote 3D LiDAR point cloud map as $P\in{\mathbb{R}^{3\times{N}}}$, where $N$ represents the number of 3D points in the map.  $T=(R, \bf t)$ represent the 6-DoF pose, where $R\in SO(3)$ is the rotation matrix and $\bf t\in {\mathbb{R}^{3}} $ is the translation vector.
% Let $K\in{\mathbb{R}^{3\times3}}$ represent the camera intrinsic, $M\in{\mathbb{R}^{1\times{H}\times{W}}}$ represent the LiDAR projected LiDAR depth, $P_i = (X_i, Y_i, Z_i)$ represent the single 3D point and $p_i=(u_i, v_i)$ represent the single pixel. Let $F, H, L$ represent the feature map, the heat feature and the flow feature respectively.

% 融合第一段和第二段的内容融合，写一段作用，紧跟着解释字母含义
% 暂时放在开头，这一段似乎不能放在 IIIB, 因为 IIIB的目标是压缩地图而非 locate
As depicted in Fig. \ref{figure:pipeline}, the proposed LHMap-loc pipeline aims to locate the monocular camera image $I\in{\mathbb{R}^{3\times{h}\times{w}}}$ within the pre-built LiDAR point cloud map $P\in{\mathbb{R}^{3\times{n}}}$, where $h$ and $w$ represent the height and width of the image, respectively, and $n$ represents the number of 3D points in the map. In this pipeline, we realize cross-modal monocular localization through two main procedures: the offline LHMap construction procedure and the online pose regression procedure.
Regarding the offline LHMap construction procedure, we feed the pre-built dense point cloud map $P$ and offline camera images into the network to construct the LHMap. During this procedure, the dense point cloud map is compressed while maintaining the key features used for localization. This procedure is described in detail in Sec. \ref{section:Offline Map Generation Network}. As for online pose regression, the LHMap and online RGB images are fed into an end-to-end network to regress the 6-DoF poses $q\in \mathbb{H}, t \in \mathbb{R}^{3}$, where $q$ represents the quaternion and $t$ is the translation vector. We achieve real-time cross-modal localization based on 2D flow feature embedding and spatial attention weighting. This procedure is detailed in Sec. \ref{section:Pose regression network}.

\subsection{Offline LHMap Generation Network}
\label{section:Offline Map Generation Network}

%直接讲具体针对输入的操作
As shown in Fig. \ref{figure:pipeline}, we use the offline LHMap generation network to compress the pre-built LiDAR point cloud map. It is realized through two stages. 

% offline包括两个stage，ref fig2, 第一个stage，第二个stage.
In the first stage, we realize the point selection to compress the dense map and pose supervision to refine the generated local map.
To satisfy the requirement for point selection and map compression, the projected LiDAR depth $D_{gt}$ is used to construct an evaluation system for point clouds. It is calculated as:
\begin{equation}
     D_{gt}^{u, v} = z^{gt},
\end{equation}
\begin{equation}
    (u, v, 1)^T  = K\cdot(x^{gt}, y^{gt}, z^{gt}, 1)^T =K\cdot{T_{gt}^{-1}\cdot(x, y, z, 1)^T}.
\label{projection_gt}
\end{equation}
Here, $(x, y, z)\in P$, $K$ represents the camera intrinsics, and $T_{gt} \in SE(3)$ represents the ground truth camera pose at each frame.
Additionally, the offline RGB image $I_{offline}$ and projected LiDAR depth $D_{init}$ are used to perform pose supervision.
Based on the initial rough camera pose $T_{init} \in SE(3) $ at each frame, which can be acquired by GPS or visual odometry, $D_{init}$ is calculated as:
\begin{equation}
     D_{init}^{u, v} = z^{init},
\end{equation}
\begin{equation}
(u, v, 1)^T  = K\cdot(x^{init}, y^{init}, z^{init}, 1)^T = K\cdot{T_{init}^{-1}(x, y, z, 1)^T}.
\label{projection}
\end{equation}
Here, $(x, y, z)^T\in P$.
Both $D_{gt}$ and $D_{init}$ contain only the depth information of point clouds. 

Firstly, feature maps $F_{I}, F_{D1}$, and $ F_{D2} $ with different scales are extracted from $I_{offline}$, $D_{gt}$, and $D_{init}$ respectively, through convolutional neural networks (CNN). $F_{D1}$, the CNN feature of the projected LiDAR depth $D_{gt}$ is used to generate the heat feature $H_{c}$. 
The point clouds are selected by evaluating heat value. Heat value is calculated by heat feature $H_c$ which is generated by $F_{D1}$. Each element $h_k^{i,j}\in H_{c} (i\in \{1,2,..., h\}, j \in \{1,2,..., w\}$) is used to calculate heat value $h^{i,j}$ for point clouds evaluation as:  

% The $D_{init}$ feature maps $F_{l2}$ and the RGB image feature maps $F_{i}$ are used to calculate optical flow embedding $E_{l1}$ based on the iterative optimization structure from PWCNet\cite{sun2018pwc}.
% 然后，我们将用真值位姿 qt 对xx 特征的生成进行监督
%突出重点，后续
% The heat feature $H_{c}$ is not only used to generate the coarse local LHMap, 
% %生成最终的Coarse LOCAL LHMap
% but also participates in pose supervision.
%参与Pose Supervision

\begin{equation}
    h^{i,j} = Mask^{i,j} \cdot \sum_{k=1}^C h_{k}^{i,j},
\end{equation}

\begin{equation}
Mask^{i,j}= \left \{
\begin{array}{ll}
    0,    & M_{gt}^{i,j} = 0\\
    1,    & M_{gt}^{i,j} \neq 0\\
\end{array}.
\right.
\end{equation}
Here, $C$ represents the number of channels of $H_{c}$.
% k top select 算法的公式实现
Subsequently, points exhibiting the highest heat values are selected to constitute the coarse local LHMap, denoted as 
$M_c$.
\begin{equation}
        M_c^{i,j} = TopN(h^{i,j}), 
\end{equation}
\begin{equation}
    TopN(h^{i,j}) =  \left \{
    \begin{array}{ll}
    D_{gt}^{i,j},  & if\ h^{i,j}\ ranking\ top\ N \\
    0,             & others \\
\end{array}.
\right.
\end{equation}
During the generation of $M_c$, the pose supervision is adopted to guide the procedure. The pose supervision module incorporates two inputs: the heat feature $H_c$, and the optical flow embedding $E_{D}$, which is derived  from $F_{D2}$ and $F_I$ based on the iterative optimization structure of PWCNet\cite{sun2018pwc}. Pose supervision is realized by pose calculation module, detailed in Sec. \ref{section:Pose regression network}. 

The single stage 1 learning fails to converge. Therefore, we propose the second stage to refine LHMap. 
In the second stage, we apply $\Delta T = T_{init}^{-1}\cdot T_{gt}$ to the coarse local LHMap $M_c$ to recover the initial localization results.
% corresponding
The initial coarse local LHMap $M_c^{init}$ and the offline RGB image $I_{offline}$ are used for further pose supervision.  
Because both stages share the same offline RGB image, they share the same feature maps $F_{I} $ of the RGB image naturally, while the feature maps $F_{M}$ of the initial coarse local LHMap $M_c^{init}$ are regenerated. Then, the heat feature $H_M$ is generated by $F_M$ and the flow embedding $E_M$ is generated by $F_{M}$ and $F_{I}$. At last, they work together for pose supervision.
Pose supervision is realized by the pose calculation module which is introduced in Sec. \ref{section:Pose regression network}. 
In this stage, we regress another set of 6-DoF pose $q_1, t_1$. Both $q_0, t_0$ and $q_1, t_1$ refine the local LHMap by optimising the heat feature $H_c$.

The output of this network is the LiDAR point cloud Heat Map (LHMap) combined by the refined local LHMap at each frame. Though the local LHMap contains only the depth information, by taking the inverse of the projection formulation, we can obtain the 3D coordinates information $P_k$ at each frame $k$. With the knowledge of the ground truth camera pose $^{w}T_{k}$ at frame $k$ and the points $P_k$ of frame $k$, we can convert $P_k$ to the world frame:
\begin{equation}
^wP_k = {^wT_{k}} \cdot  P_k .       
\label{i to world}
\end{equation}
Here, $^wP_k$ represents the points at the frame $k$ in the world coordinate system. The LHMap is constructed by uniting all the points $^wP_k$ together through an union operation $\cup_k$:
\begin{equation}
LHMap = \cup_k{^wP_k}.    
\label{t to world}
\end{equation}
%组合方式的详细解释

The loss function of the offline heat map generation network is similar to CMRNet\cite{cmrnet}. 
Let $q_{gt}$ and $t_{gt}$ represent the ground truth camera pose. The angular distance $L_q$ between quaternions is used to evaluate the rotation loss. The L1-smooth loss $L_t$ is used to evaluate the translation loss, which is defined as:
\begin{equation}
    \mathcal{L}_q(q,\ q_{gt}) = D(q \bigotimes {inv(q_{gt})}),
\end{equation}
\begin{equation}
    D(q) = \arctan((\sqrt{b^2+c^2+d^2},\ |a|)),
\end{equation}
\begin{equation}
    \mathcal{L}_t(t,\ t_{gt}) = L_1smooth(t-t_{gt}),
\end{equation}
Here, $ \{a, b, c, d\}$ are the components of quaternion $q$ and $\bigotimes$ is the multiplicative operation between two quaternions.
The pose loss is defined as:
\begin{equation}
\mathcal{L}_p = \mathcal{L}_t+\lambda\mathcal{L}_q, \lambda \geq 1          
\label{lamLoss}.
\end{equation}
The pose $t_0, q_0$ regressed by the pose supervision module in stage 1 and the pose $t_1, q_1$ regressed by the pose supervision module in stage 2 are both taken into account for better supervision. Therefore, the total loss is defined as:
\begin{equation}
\mathcal{LOSS}_1 = \alpha\mathcal{L}_{p0}+\beta\mathcal{L}_{p1}, \ \alpha + \beta = 1.
\label{loss1}.
\end{equation}

%-----------------------------------------------------------------

\subsection{Online Pose Regression Network}
\label{section:Pose regression network}
%分级的问题，文字体现
This network is used for real-time monocular localization. The inputs are the online RGB image $I_{online}$ and the real-time LHMap $M^r$. The $M^r$ is constructed by projecting $^{t}P$ at each local LHMap stored in the first network to the image plane according to the function in (\ref{projection}). 
% Both inputs are fed into the pose regression module.
% The main purpose of the module is regressing 6-DoF pose $\Delta H$ end-to-end.

Firstly, feature maps $F_{I}^r$ and $F_{M}^r$ are extracted from both inputs $I_{online}$ and real-time local LHMap $M^r$ through convolutional neural networks (CNN). 

Then, the feature maps $F_{M}^r$ are used to calculate the 2D flow embedding $E_M^r$ along with the RGB image feature maps $F_{I}^r$ and to generate the heat feature $H_{M}^r$ alone. $E_M^r$ here is calculated the same as PWCNet\cite{sun2018pwc}. The usage of the heat feature $H_{M}^r$ enables the pose regression to focus on effective features. Therefore, the supervision of the 2D flow embedding $E_M^r$ and the regression of 6-DoF pose can achieve better performance. The cost volume $V$ is then calculated by feeding $H_{M}^r$ into the softmax layer to generate the coefficients and multiplying the coefficients with $E_{M}^r$. The cost volume $V$ is calculated as:
\begin{equation}
    \mathcal V = \sum_{h\times w}{E_{M}^r \odot  Softmax_{h\times w}(H_{M}^r)},
    \label{}
\end{equation}
where $\odot$ means element-wise product, $softmax_{h\times w}$ means apply softmax to height and width dimensions of $H_{M}^r$.

\begin{figure}[t]
   \centering
   \includegraphics[width=1\linewidth]{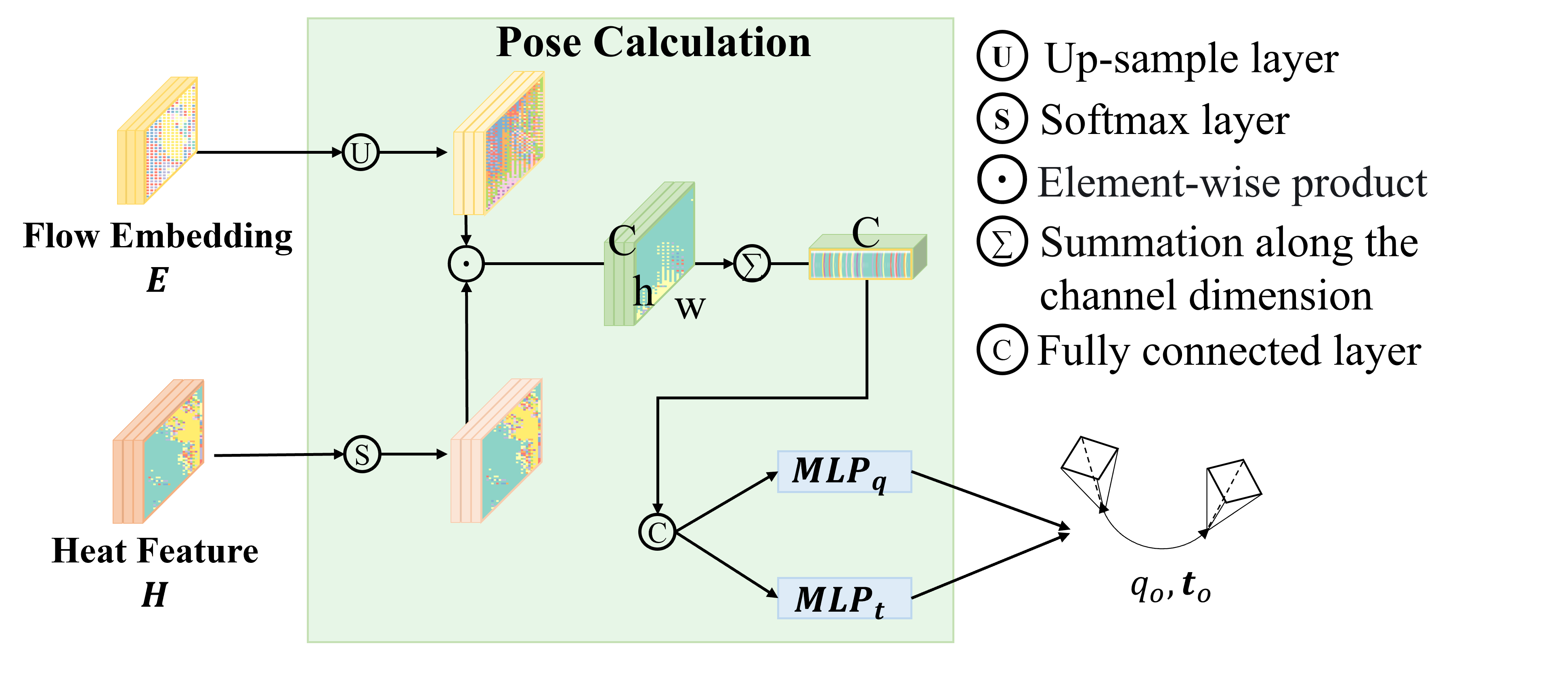}
   %\vspace{-0.2cm}
   \caption{The details of the regression part. Multiply flow embedding $E$ and up-sampled heat feature $H$ as inputs, and then calculate weighted features. The result is fed into $MLP_q$ and $MLP_t$ to regress 6-DoF poses}
   \vspace{-0.3cm}
   \label{figure:regression}
\end{figure}

At last, the cost volume is fed into separate MLPs for pose regression:
\begin{equation}
   q_o = MLP_q(V),\ \ t_o =  MLP_t(V).
    \label{}
\end{equation}
The pose regression is realized by the pose calculation module as shown in Fig. \ref{figure:regression}. The resolution of the flow feature may be different from that of the heat feature. Therefore, the flow feature is transferred to the up-sampled layers to maintain the same resolution as the heat feature before being multiplied with it. The multiplication result then accumulates all the elements across the height and width dimensions before being fed into fully connected layers, which are denoted as  $MLP_q$ and  $MLP_t$. The outputs of this network are 6-DoF poses $t_o,q_o$.

The loss function used here follows \cite{li2019net}. Adding two trainable parameters $w_x$ and $w_q$, the loss function is defined as:
\begin{equation}
    \mathcal{LOSS}_2 = e^{-w_x}\mathcal{L}_t+w_x+e^{-w_q}\mathcal{L}_q+w_q.
    \label{baseLoss}
\end{equation}

% The LHMap generation module and the pose regression module work separately. Their structures and parameters are easily adjusted and modified independently.
% They can also be replaced easily by other point selection methods or pose regression methods. As a result, our methods are flexible and extendable.
%-----------------------------------------------------------------
% 字体，公式斜体 右上角

\subsection{Training Details}
\label{section:Training Details}
We implement our network using PyTorch. For the offline heat map generation network, it is trained for 120 epochs using the ADAM optimizer, with a batch size of 8 and a learning rate of 1e-4. We apply the loss function as defined in (\ref{loss1}), setting the parameters to $\lambda=10$, $\alpha=0.6$, and $\beta=0.4$. The top 5000 point clouds are selected for storage and further processing. For the online pose regression network, it is trained for 150 epochs, utilizing the Adam optimizer \cite{kingma2014adam} with a batch size of 12 and a learning rate of 1e-4. The loss function (\ref{baseLoss}) is employed with the initial values set to $w_x=0$ and $w_q=-2.5$. All training and evaluation activities are performed on a single NVIDIA GTX 2080 Ti.

% \begin{figure*}[h]
%   \centering
%   \includegraphics[width=0.8\linewidth]{examples_new.png}
%   \caption{Some visualization results of KITTI sequence 00 in the monocular localization. Left from up to down: LiDAR point cloud map; Original dense LiDAR-image; RGB image. Right from down to up: Heatmap used for generating LHMap, Refined Local LHMap; LHMap. }   
%   %\vspace{-0.2cm}
%   \label{figure:sample}
% \end{figure*}

\section{Experiments}

\begin{figure*}[h]
   \centering
   \includegraphics[width=0.95\linewidth]{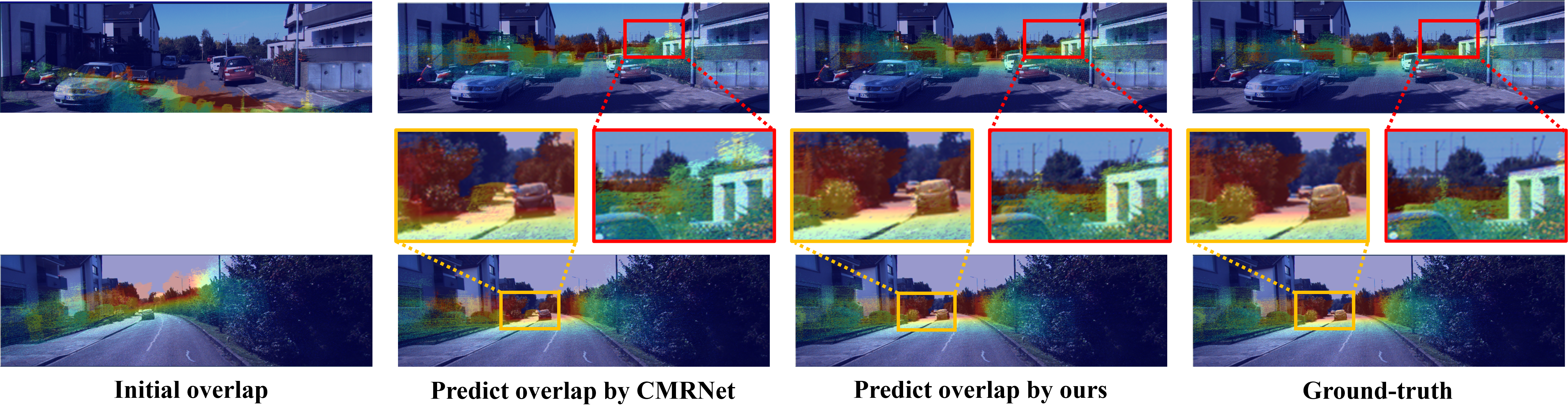}
   \vspace{-3mm}
   \caption{Qualitative results of LiDAR-image registration on KITTI\cite{kitti} dataset.}   
   %\vspace{-0.2cm}
   \label{figure:overlap}
\end{figure*}
% --------------------------------------------------------------------------------------

\setlength{\tabcolsep}{2.5mm}
\begin{table*}[h]
\centering
\caption{Quantitative localization results of training and testing on single KITTI\cite{kitti} and  Argoverse\cite{argo} dataset}
\scalebox{1}{  
    \begin{tabular}{lc|cc|cc|cc|cc}
    \hlineB{3}
    \multicolumn{1}{c}{}  
      
    & &\multicolumn{2}{c|}{CMRNet\cite{cmrnet}} 
    % &\multicolumn{2}{c}{CMRNet++\cite{cmrnet++}} 
    & \multicolumn{2}{c|} {HyperMap\cite{hyper}} &\multicolumn{2}{c|}{PosesAsQueires\cite{PosesAsQueires}}  
    & \multicolumn{2}{c}{Ours}\\
    & & Transl.[m] & Rot.[deg]  
    % & Transl.[m] & Rot.[deg] 
    & Transl.[m] & Rot.[deg]   
    & Transl.[m] & Rot.[deg]  
    & Transl.[m] & Rot.[deg]  \\  \hline
    KITTI\cite{kitti} 
    &Iter1   
    & 0.51 & 1.39 
    % & 0.55(2.18)  & 1.46 
    &0.48 &1.42  
    & 0.41 & 1.39   
    &\bf0.21 & \bf0.94   \\
    &Iter2  
    & 0.31  & 1.09 
    % & 0.22  & 0.77    
    &- &-    
    & 0.20 & 0.90  
    & \bf 0.06 & \bf 0.35  \\
    &Iter3  
    & 0.27  & 1.07  
    % & 0.14  & 0.43   
    &- &-    
    & 0.20  & 0.79  
    & \bf 0.03 & \bf 0.33      \\ \hline
    Argoverse\cite{argo} 
    &Iter1  
    &0.90 &1.78 
    % &0.80(6.24) &1.55  
    &0.58 &\bf0.93   
    &- &-   
    &\bf0.20 &0.99    \\
    &Iter2   
    & 0.80    & 1.56
    % &0.34 &\bf0.58     
    &- &-    
    & - & -   
    &\bf0.10  &\bf0.66  \\
    &Iter3  
    & 0.67   & 1.52
    % &0.25 &\bf0.45     
    &- &-    
    & -  & -   
    &\bf0.09  &\bf0.57       \\ \hlineB{2.5}
    
    \end{tabular}
}

\label{table:Results}
\end{table*}

% ------------------------------------------------------------------------------------------------------------------

% ---------------------------------------------------------------------------------------------------------

\setlength{\tabcolsep}{3.0mm}
\begin{table*}[h]
\centering
\caption{Quantitative results of methods using sharing weights on KITTI\cite{kitti} and Argoverse\cite{argo} dataset }
\scalebox{0.85}{

\begin{tabular}{lc|ccc|ccc|ccc}
\hlineB{3}
& 
&\multicolumn{3}{c|}{CMRNet++\cite{cmrnet++}} &\multicolumn{3}{c|}{I2D-Loc\cite{i2d}} 
& \multicolumn{3}{c}{Ours}\\
& 
& Transl.[m] & Rot.[deg] & failure rate[\%] 
& Transl.[m] & Rot.[deg] & failure rate[\%]
& Transl.[m] & Rot.[deg] & failure rate[\%]
 \\ \hline
& KITTI\cite{kitti} 
& 0.55 & 1.46 & 2.18 
& \bf0.17 & \bf0.70 & 1.61  
& 0.26 & 1.50 & \bf0   
\\
& Argoverse\cite{argo} 
& 0.80 & 1.55 & 6.24 
& 0.47 & \bf0.71  & 7.52 
& \bf0.29 & 1.69 & \bf0
\\ \hlineB{2.5}

\end{tabular}

}

\label{table:norm}
\end{table*}

% -----------------------------------------------------------------------------------------------------------
\linespread{0.9}
\setlength{\tabcolsep}{2.0pt}
\begin{table}[h]
\centering
\caption{Ablation experiments on pose regression and Top N selection.}
\begin{center}
\vspace{-3mm}
% \resizebox{1\columnwidth}{!}{

\begin{tabular}{lcc|ccc|cccccc}
\hlineB{3}
& \multicolumn{2}{c|}{Method}
& \multicolumn{3}{c|}{Top N}
& \multicolumn{2}{c}{Transl.[m]} 
& \multicolumn{2}{c}{Rot.[deg]} \\ 
&CMRNet &ours &N = all & N = 10k & N = 5k & Mean & Median &Mean & Median \\ 
\hline
&\checkmark & &\checkmark & & & 0.57  & 0.51 
& 1.80  & 1.39  \\
&\checkmark & & & &\checkmark & 0.31
& 0.26 &1.99  & 1.83  \\
& &\checkmark &\checkmark & & & 0.64  & -  &  1.92 & -  \\
& &\checkmark & &\checkmark & & 0.28  &0.23  &  1.41 & 1.27  \\
& &\checkmark & & &\checkmark & \bf0.25  & \bf0.21  & \bf1.04 & \bf0.94  \\
% &\checkmark & & & \bf0.23  & \bf0.19  & \bf1.0499  & \bf0.9367 \\ 
\hlineB{2}
\end{tabular}
% }

\end{center}
\label{table:ablation_2}
\end{table}
\linespread{1}

%-----------------------------------------------------------------------------------------------------------

\linespread{0.9}
\setlength{\tabcolsep}{0.8mm}
\begin{table}[h]
\centering
\caption{Performance comparison of different sources of heat feture.}
\begin{center}
\vspace{-2mm}
% \resizebox{1\columnwidth}{!}{
\begin{tabular}{lccc|cccc}
\hlineB{3}
& \multicolumn{3}{c|}{Heat Map Generation}
% & \multicolumn{3}{c}{n}
& \multicolumn{2}{c}{Transl.[m]} 
& \multicolumn{2}{c}{Rot.[deg]} \\ 
& Randomly & RGB image & LiDAR depth 
% & n=all & n=10000 & n=5000
& Mean & Median &Mean & Median \\ 
\hline
&\checkmark & &  & 1.15 & - & 1.74 & -  \\
& &\checkmark &  & 0.30 & 0.25 & 1.62 & 1.32  \\
& & &\checkmark  & \bf0.25 & \bf0.21 & \bf1.04 & \bf0.94  \\
\hlineB{2}
\end{tabular}
% }

\end{center}
\label{table:ablation_1}
\vspace{-7mm}
\end{table}
\linespread{1}
% -------------------------------------------------------------------------------------------------------

% ------------------------------------------------------------------------------------------------------

\subsection{Datasets and Data Preprocessing}
\subsubsection{KITTI dataset \cite{kitti}} The LiDAR maps, the ground truth poses, and the initial poses are generated following CMRNet\cite{cmrnet} and HyperMap\cite{hyper}. KITTI Odometry sequences 03, 05, 06, 07, 08 and 09 are selected to be the training set (11426 frames) and sequence 00 is the evaluation set (4541 frames). The methods for generating LiDAR maps and the projected LiDAR depth follow the same procedures as outlined in CMRNet\cite{cmrnet}.
\subsubsection{Argoverse dataset \cite{argo}} Images from the central forward facing camera that provides $1920\times1200$ images are used for localization on Argoverse dataset. These images are down-sampled to $960\times600$ according to \cite{cmrnet++}. The ground-truth maps are built with voxel resolutions of 0.1m.
%数据集的标注形式
Sequences train1, train2, train3, train4 and val are used as the training set (17614 frames), and the test sequence is used as the evaluation set (4168 frames). Besides, following \cite{hyper}, some frames affected dynamic objects are removed from the training set.

We also apply an iterative refinement approach following \cite{cmrnet}. For the first iteration, we introduce uniformly distributed noise ranging from $[-2m, 2m]$ for translation and $[-10^\circ, 10^\circ]$ for rotation to the ground truth poses $H_{gt}$. Subsequent noise levels are set to $[-1m, 1m]$ with $[-2^\circ, 2^\circ]$ for the second iteration, and $[-0.6m, 0.6m]$ with $[-1^\circ, 1^\circ]$ for the third iteration.

% To simulate the initial pose, we also applied uniform random noise which ranged from $[-2m, 2m]$ for translation and $[-10^\circ, 10^\circ]$ for rotation to the ground truth pose $H_{gt}$ for each sample independently followed by \cite{cmrnet, cmrnet++,hyper,i2d}. The noises for second and third iteration are $[-1m, 1m]$, $[-2^\circ, 2^\circ]$ and $[-0.6m, 0.6m]$, $[-1^\circ, 1^\circ]$ respectively.

%-------------------------------------------------------------------------
\subsection{Performance}

% 图表前的空格，分析
As for qualitative results, Fig. \ref{figure:overlap} exhibits the results of LiDAR-image registration using predicted poses between camera images and projected LiDAR depths. Compared with \cite{cmrnet}, our method can achieve better LiDAR-image registration with overlap patterns more similar to the ground-truth. Moreover, in regions with sparse features, our pipeline can also achieve robust pose regression thanks to the pre-built LHMap and the spatial attention weighting algorithm. 

Table \ref{table:Results} shows the quantitative  monocular localization results of different methods under 3 iterations. For a fair comparison, all methods listed in the table follow the same selection of the training set and the test set. With regard to the KITTI dataset and Argoverse dataset, our pipeline enables more accurate monocular localization evaluated by both translation and rotation errors by a large margin compared with the SOTA methods: CMRNet\cite{cmrnet}, 
HyperMap\cite{hyper}, and PosesAsQueires\cite{PosesAsQueires}. Moreover, one iteration of our method yields even higher localization accuracy than three iterations of CMRNet. In addition, our method achieves a more significant accuracy improvement during iterative optimization.

% It is worth noting that our failure rate is zero both in KIITI\cite{kitti} and Argoverse\cite{argo} datasets, which means our network is more robust. 

\begin{figure}[t]
   \centering
   \includegraphics[width=1\linewidth]{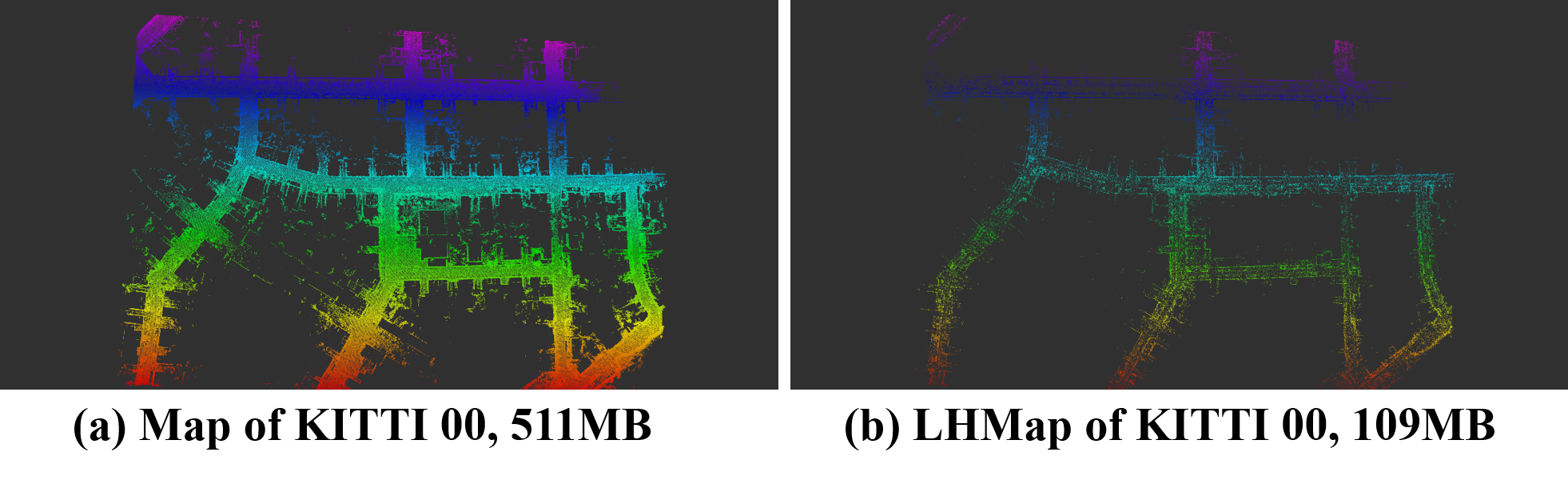}

   \vspace{-0.2cm}
   \caption{3D point cloud map of KITTI sequence 00. (a) Original LiDAR point cloud map. (b) LHMap.}
   \label{figure:map}
\end{figure}

Besides, Fig. \ref{figure:map} displays the visualisation results of the point cloud map on KITTI sequence 00 with the voxel size equals to 0.1m. As shown in Fig. \ref{figure:map}, the generated LHMap retains the main structural features of the point cloud. Compared with the original map, our LHMap compresses the map by 80\%. Meanwhile, LHMap achieves better performance for monocular localization due to its effective feature extraction. 

We also conduct experiments on projecting the RGB image and the projected LiDAR depth to the normalization plane, which aims to achieve mapping and monocular localization with different cameras. In this procedure, we first convert every pixel in the RGB image and every point in the point cloud map to the normalization frame. Then we map $\{(x,y)|-0.8<x<0.8, -0.4<y<0.4\}$ in the normalization coordinate to the image plane $\{(u,v)|0\leq u<768, 0\leq v<384\}$.
Then, we train the pipeline on KITTI sequences 03, 05, 06, 07, 08, 09 and Argoverse sequences Train1, Train2, Train3, and test on KITTI sequence 00 and Argoverse sequence Train4 following \cite{i2d}. The results are shown in Table \ref{table:norm}. According to CMRNet++\cite{cmrnet++}, we define the situation as a failure if the translation error is larger than 4m. The results demonstrate that our methods achieve competitive localization accuracy especially on Argoverse. Remarkably, our failure rate is zero on both KITTI and Argoverse. It is demonstrated that our method is more robust to challenging environments such as rural roads lacking structured geometry features.
% -----------------------------------------------------------------------------------------------

% ----------------------------------------------------------------------------------------------------------------------------------------------------------------

\subsection{Ablation Study}
To verify the contribution of key modules in our LHMap-loc pipeline, a series of ablation experiments are designed. The experiments are mainly carried out from three aspects: the strategy of heat map generation, the density of LHMap and the method of pose regression. We thoroughly evaluate all methods on the KITTI dataset\cite{kitti} with one iteration. And the results are as shown in Table \ref{table:ablation_2} and Table \ref{table:ablation_1}. 

First of all, in Table \ref{table:ablation_2} different pose regression modules are tested  in the online pose regression network using different $Top N$ mapping points per frame. We select $N = all$, $N=10000$ and $N=5000$ points separately and $N = 5000$ achieves even better localization results than the other two cases. The pose regression module is also replaced by CMRNet, and the results demonstrate that our improvements are effective.

Additionally, we test on different ways to generate LHMap. In Table \ref{table:ablation_1}, heat features generated by the RGB image are leveraged  instead of the projected LiDAR depth to construct LHMap. Offline maps are also generated by random selection. The experiments demonstrate that our LHMap generation strategy achieves better localization accuracy. 
%取得更好的定位精度

\begin{table}[t]
  \centering
  \caption{comparison of Time consumption in each method}
  \begin{tabular}{ccccccc}
  \hlineB{3}
  \multicolumn{1}{c}{Time/ms}      & CMRNet  &CMRNet++ & I2D-loc &PosesAsQueries &  Ours   \\ \hline
  Pre-process Time      &101.890   &- & - & - & \bf1.844  \\
  Inference Time    & 6.868 & 1250. & 80. & - & \bf11.079 \\
  Total   & 109.910 & $>$1250. & $>$80. &14.925 & \bf13.402 \\ \hlineB{2}
  \end{tabular}
   \vspace{-3mm}
\label{table:time}
\end{table}
\subsection{Time Consumption}
Localization has a strict requirement for real-time performance, therefore we test the time consumption in the process of pose regression. The pre-process time, which includes the time for data loading, projecting and occlusion filtering and inference time are evaluated and displayed in Table \ref{table:time}. Every sample is tested and averaged on KITTI sequence 00 with batch size equals to 1. The results demonstrate our network can perform pose regression at about 77 frames per second (FPS), while CMRNet spends much more time in pre-processing data. In conclusion, our methods spend much less time in localization than \cite{cmrnet}, \cite{cmrnet++}, \cite{i2d} and \cite{PosesAsQueires}.

\subsection{Real-world Experiments}

\begin{figure}[t]
   \centering
   \includegraphics[width=1\linewidth]{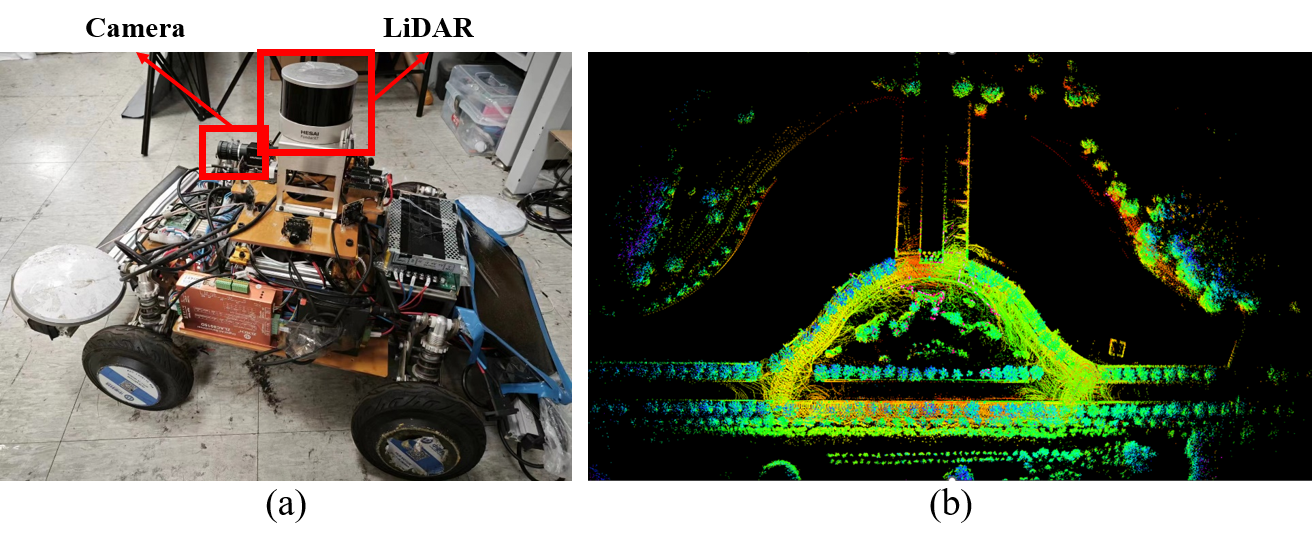}
   \vspace{-8mm}
   \caption{(a) The wheeled vehicle for SJTU-dataset collection. \ \ (b) One scenario of SJTU-dataset.}
   \label{figure:vehicle}
\end{figure}

To validate our methods in real-world scenarios, we collect data using a Hesai PandarXT-16 LiDAR and an industrial camera (MV-CA013-21UC) at Shanghai Jiao Tong University (SJTU), specifically around the lake and the administration building. The data is further processed by FAST-LIO\cite{xu2022fast} and RTK to acquire ground truth poses and the pre-built point cloud map. It is noteworthy that the collected scenarios are challenging. Because these scenarios are primarily composed of trees and shrubbery, while lacking the buildings like KITTI and Argoverse. Meanwhile, the equipment is carried on a low-speed, remote-controlled vehicle, as shown in Fig. \ref{figure:vehicle}, which means the field of view of the LiDAR and camera is entirely different from that of KITTI and Argoverse. In the SJTU dataset, our methods also achieve extremely accurate localization results, as shown in Table \ref{table:SJTU}, not affected by changing scenarios. Overall, our methods demonstrate robustness in facing the challenges of real-world scenarios.
% ---------------------------------------------------------------------------
% --------------------------------------------------------------------------------

\setlength{\tabcolsep}{1.5mm}
\begin{table}[t]
  \centering
  \caption{Localization results in SJTU-dataset}
\scalebox{1}{  
  \begin{tabular}{ccccccc}
  \hlineB{3}
  &\multicolumn{2}{c}{Transl.[m]}
  &\multicolumn{2}{c}{Rot.[deg]}
  & \multirow{2}{*} {Map size}\\ 
  & Mean & Median & Mean & Median \\ \hline
  
  CMRNet(All Points)
  & 0.95 & 0.86 & 1.66 & 1.42  & 197.99MB \\
 Ours (All Points)  
  & 1.02 & 0.95 & \bf0.72 & \bf0.64  & 197.99MB \\ 
Ours (5k Points)
  & \bf0.22 & \bf0.19 & 1.01 & 0.85  & \bf39.51MB \\ \hlineB{2}
  \end{tabular}
}  
\label{table:SJTU}
   \vspace{-3mm}
\end{table}

%------------------------------------------------------------------------
\section{Conclusion}
This paper uses offline heat map generation network to construct LHMap. Further, online pose regression is realized by an end-to-end pose regression network for LHMap and real-time RGB images. Through extensive experimental results, the effectiveness of LHMap is demonstrated in improving localization accuracy and reducing the size of LiDAR maps. Overall, to our knowledge, the proposed LHMap-loc in this paper achieves higher accuracy and is more robust than SOTA learning-based monocular localization.

%\addtolength{\textheight}{-0cm}   % This command serves to balance the column lengths
                                  % on the last page of the document manually. It shortens
                                  % the textheight of the last page by a suitable amount.
                                  % This command does not take effect until the next page
                                  % so it should come on the page before the last. Make
                                  % sure that you do not shorten the textheight too much.

\bibliographystyle{IEEEtran}  % set style to IEEE
\bibliography{IEEEabrv,root} % set reference file name

\end{document}